\documentclass[mlabstract,onecolumn]{jmlr}

\jmlrproceedings{}{}
\firstpageno{1}
\author{James Hazelden$^{1,2}$\thanks{Accepted at NeurReps 2026, NeurIPS Workshop on Symmetry and Geometry in Neural Representations}\\
\textnormal{$^1$ Applied Mathematics, University of Washington\\
$^2$ The Allen Institute\\
\text{  } Seattle, WA, USA}\\
\texttt{jhazelde@uw.edu}}
\title[Green's Operator for Multi-Task Computation]{Disentangling Computation in Multi-Task Neural Networks with the Green's Operator}

\newcommand{\opP}{\mathrm{P}}

\newcommand{\ip}[2]{\left\langle #1,#2\right\rangle}
\newcommand{\norm}[1]{\left\lVert #1\right\rVert}

\newcommand{\graphic}[2]{%
\includegraphics[width=\linewidth,height=#2,keepaspectratio]{figures/#1}}

\begin{document}
\maketitle
\vspace{-2em}

\begin{abstract}
How is computation organized and reused across tasks and time in a trained recurrent network?
Most analyses emphasize the geometry of neural activity, dynamical motifs, or local perturbation growth.
We instead study the network's global first-order perturbation response.
The finite-horizon Green's operator maps perturbations at each source along a trajectory to their downstream state-space responses and therefore directly represents perturbation routing.
Simple reductions of this operator provide task-to-task and time-to-time views of the same computation, while matrix-free products make these views accessible without constructing the full operator.
In a flexible multitask recurrent network, task reductions reveal structured reuse of known computational motifs, while temporal reductions reveal causal pathways and how they emerge during training.
Our main point is simple: the Green's operator provides a global response geometry for mapping the organization of learned dynamical computation.
\end{abstract}


\section{Introduction}
A recurrent network can solve many tasks with the same state space and parameters.
This raises a basic question: \emph{how is computation organized and reused across tasks and time?}
Flexible recurrent networks can reuse attractors and other dynamical motifs across related computations~\citep{sussillo2013opening,turner2023simplicity,driscoll2024flexible}.
The challenge is to obtain a global view of this organization without separately cataloguing every trajectory, fixed point, task condition, time interval, and perturbation direction.

Activity-based analyses describe where trajectories lie and how representations co-vary.
Lyapunov analyses describe how perturbations grow or decay and along which directions~\citep{engelken2020lyapunovspectrachaoticrecurrent,vogt2022lyapunov,storm2024finite}.
Neither representation is organized directly by the question we ask here: \emph{if a perturbation enters at this point in the computation, where does its influence appear later, and is that response reused in another task?}

We study this source-to-destination structure through the finite-horizon Green's operator, denoted $\opP$.
For a trajectory of a dynamical model, $\opP$ is the linear map from perturbations applied across the computation to the resulting first-order state perturbations.
Its blocks encode causal response between particular source and destination points.
The same global object is naturally indexed by task, time, trial, and hidden state: reducing different axes asks either \emph{which computations are reused?} or \emph{when is influence routed?}
Our contribution is intentionally narrow: we isolate why this routing is distinct from activity and local spectral stability, then use task- and time-level reductions to map computation in a well-studied flexible multitask RNN~\citep{driscoll2024flexible}.

\section{The Green's operator: a global response map}
\label{sec:green}
Consider a recurrent system linearized along a trajectory,
\begin{equation}
    \delta h_{t+1}=J_t\,\delta h_t+u_t,
    \label{eq:linrec}
\end{equation}
where $u_t$ is an injected perturbation.
Stacking perturbations and downstream state responses over a finite horizon gives
\begin{equation}
    \delta h = \opP u.
\end{equation}
For $t>s$, the corresponding block is
\begin{equation}
    \opP_{t,s}=J_{t-1}J_{t-2}\cdots J_s,
    \label{eq:block}
\end{equation}
with the identity on the diagonal under the convention used here.
Thus $\opP_{t,s}$ directly maps a perturbation at source time $s$ to its first-order effect at destination time $t$.
The equivalent global implicit definition $\opP=[D_hF]^{-1}$ and its path expansion are given in Appendix~\ref{app:operator}.

\begin{figure}[t]
\floatconts{fig:toy}
{\caption{\textbf{Routing is not determined by activity or local spectrum.}
Two linear recurrent systems generate the same unperturbed trajectory and have the same eigenvalues, but differ in whether a perturbation in $y$ can influence $x$.
The finite-horizon Green's operator records this source-to-destination difference directly.}}
{\graphic{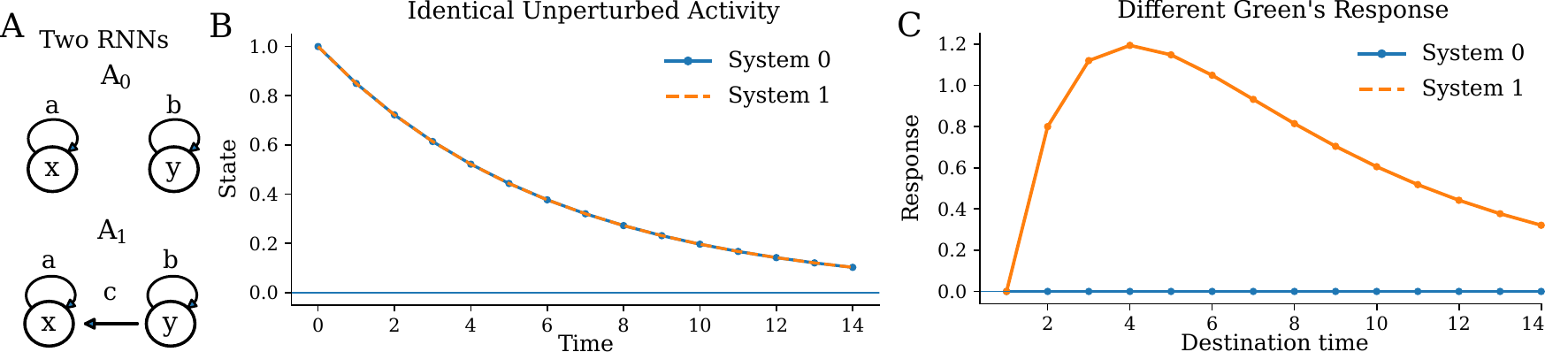}{1.35in}}
\end{figure}

For a task-conditioned trajectory $\tau$, let $\opP_\tau$ denote the corresponding response operator.
We use simple reductions rather than forming its full task/trial/time/state tensor.
A task-level representation $G_\tau=\mathcal{R}_{\mathrm{task}}(\opP_\tau)$ removes nuisance axes while retaining response geometry for task comparison.
A time-level reduction $R_\tau(t,s)=\mathcal{R}_{\mathrm{time}}(\opP_\tau)_{t,s}$ instead retains source and target time, giving a task-specific map of temporal dependence.
The full matrix is never required: a forward recurrence computes $\opP u$, a reverse recurrence computes $\opP^*v$, and these products support randomized low-rank and reduced summaries (Appendix~\ref{app:matrixfree}).

\begin{figure}[t]
\floatconts{fig:taskreuse}
{\caption{\textbf{Task-level reductions of the Green's operator expose computational reuse.}
\textbf{A}, reducing the full task/time/unit response operator gives a task-level view of perturbation reuse.
\textbf{B--E}, known motif organization, pairwise Green-response similarity, strongest inferred relations, and a low-dimensional embedding provide complementary views of task organization.
\textbf{F}, Green-response similarity and hidden-state covariance are related but non-equivalent; points distinguish task pairs from the same versus different families.}}
{\graphic{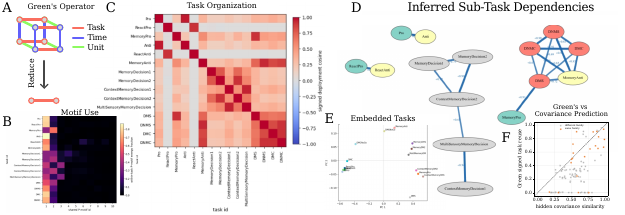}{2.45in}}
\end{figure}

\section{A minimal example: same activity, same spectrum, different routing}
\label{sec:toy}
The purpose of the toy is not to show that Lyapunov analysis fails.
It isolates a simpler point: observed state trajectories and local eigenvalue spectra do not by themselves specify source-to-destination perturbation routing.
Figure~\ref{fig:toy} compares two stable triangular linear RNNs that generate exactly the same trajectory from the chosen initial condition and have the same eigenvalues, but differ by one off-diagonal route $y\rightarrow x$.
A perturbation to $y$ therefore has no downstream effect on $x$ in one system and a persistent effect in the other.
The exact response is derived in Appendix~\ref{app:toy}.

\section{Task organization in a flexible multitask RNN}
\label{sec:tasks}
We next ask: \emph{which computations are shared across tasks?}
We analyze a 256-unit leaky RNN trained jointly on the 15-task family of~\citet{driscoll2024flexible}, using the converged checkpoint at step $680{,}000$.
Prior work already identifies motif reuse in this family; we ask whether it is visible directly in perturbation-response geometry.

For each task condition $\tau$, we compute the same reduced response $G_\tau$ and compare pairs using normalized Frobenius similarity,
\begin{equation}
 S(\tau,\tau')=\frac{\ip{G_\tau}{G_{\tau'}}_F}{\norm{G_\tau}_F\norm{G_{\tau'}}_F}.
 \label{eq:cosine}
\end{equation}
The resulting matrix gives a global view of task-conditioned response reuse.
It shows block structure aligned with known sub-computations, while disagreements with hidden-state covariance show that activity geometry and response routing are related but non-equivalent.
We use the known motif organization as an external reference, not as a claim that Green similarity universally dominates activity-based summaries; controls are in Appendix~\ref{app:experimental}.

\section{Temporal organization emerges during training}
\label{sec:time}

\begin{figure}[t]
\floatconts{fig:time}
{\caption{\textbf{Task-specific temporal response pathways emerge during training.}
\textbf{A}, reducing the task-conditioned Green's operator while retaining source and target time gives a temporal routing map.
\textbf{B--E}, signed uniform-direction Green-response projections before training (step 0) and after training (step $680{,}000$) for ReactPro, MemoryPro, DMS, and ContMemDecision2.
Reactive computation remains comparatively local, while memory tasks develop delay-spanning pathways. Exact Frobenius-energy controls are given in Appendix~\ref{app:experimental}.}}
{\graphic{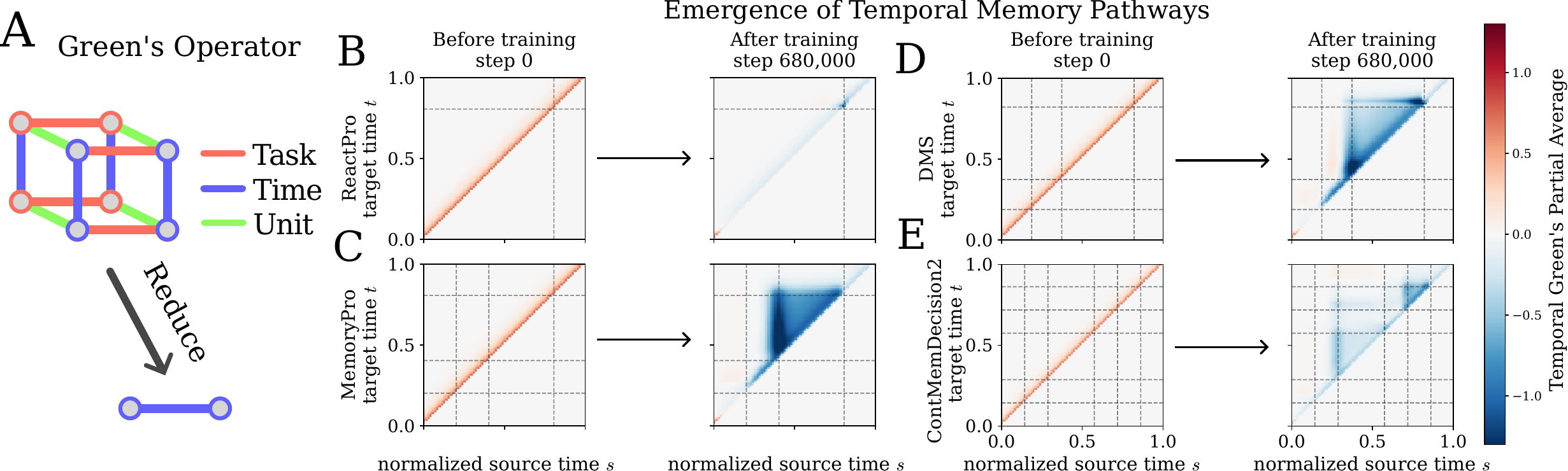}{2.35in}}
\end{figure}

The same operator asks a complementary question: \emph{when is influence routed?}
Reducing $\opP_\tau$ over trials and hidden units while retaining source time $s$ and target time $t$ gives $R_\tau(t,s)$.
Before training, response lies near the causal diagonal; after training, ReactPro remains comparatively local while MemoryPro and DMS develop delay-spanning structure and ContMemDecision2 a more restricted persistent pathway (Figure~\ref{fig:time}).

Figure~\ref{fig:time} displays a signed uniform-direction response projection, not a mathematical neuron trace.
An exact basis-invariant Frobenius reduction gives the same qualitative conclusion: MemoryPro gains substantially more long-lag response than ReactPro, while both begin near diagonal at initialization (Appendix~\ref{app:experimental}).
Training therefore organizes temporal routing differently for computations with different persistence demands.

\section{Discussion}
The task-to-task and time-to-time reductions of the Green's operator ask complementary questions: \emph{what is reused?} and \emph{when is influence routed?} This source-to-destination view complements fixed points, activity geometry, and Lyapunov analysis rather than replacing them. Because the analysis is first-order and finite-horizon and every reduction discards information, our claim is narrow: global response geometry provides a useful matrix-free map of task reuse and temporal routing.

\clearpage
\bibliography{refs}

\appendix

\section{General operator formulation}
\label{app:operator}
A broad class of explicit or implicit models can be written as a global constraint
\begin{equation}
 F(h,x,\theta)=0,
\end{equation}
where $h$ denotes the collected model state, $x$ the inputs, and $\theta$ the model parameters.
When $D_hF$ is invertible, the finite-network Green's operator is
\begin{equation}
 \opP=[D_hF(h,x,\theta)]^{-1}.
 \label{eq:implicitP}
\end{equation}
For an explicit acyclic computation graph, a convenient constraint convention gives $D_hF=I-L$, where $L$ contains one-step state dependencies and is nilpotent. Hence
\begin{equation}
 \opP=(I-L)^{-1}=I+L+L^2+\cdots+L^{K},
 \label{eq:path}
\end{equation}
for finite $K$. Each term collects paths of a given length through the computation graph. For recurrent trajectories unrolled over a finite horizon, Eq.~\eqref{eq:path} is equivalent to the block products in Eq.~\eqref{eq:block}.

This formulation also makes the connection to learning explicit. Differentiating the constraint with respect to parameters gives
\begin{equation}
 D_\theta h=-\opP\,D_\theta F,
\end{equation}
up to the sign convention used to define $F$. Thus parameter-to-state learning operators are parameter-selected sketches of the global response geometry. Related empirical operator factorizations are studied in~\citet{hazelden2026globalempiricalntkselfreferential}.

\section{Matrix-free products and randomized summaries}
\label{app:matrixfree}
For the recurrent system in Eq.~\eqref{eq:linrec}, a product $y=\opP u$ is computed by a single forward recurrence:
\begin{algorithm}[htbp]
\caption{Forward Green product $y=\opP u$}
\begin{enumerate}
\item Set $y_0=u_0$ under the identity-diagonal convention.
\item For $t=0,\ldots,T-2$, set $y_{t+1}=J_t y_t+u_{t+1}$.
\item Return the stacked response $y=(y_0,\ldots,y_{T-1})$.
\end{enumerate}
\end{algorithm}
The adjoint product $z=\opP^*v$ is computed by the reverse recurrence $z_{T-1}=v_{T-1}$ and $z_t=v_t+J_t^*z_{t+1}$.
The cost is therefore linear in the horizon times the cost of Jacobian-vector or vector-Jacobian products, without storing the full $T^2N^2$ block operator.
These products are sufficient for randomized range finding and SVD, Gram products, and stochastic reduced traces or block-energy estimators.

\section{Toy derivation}
\label{app:toy}
For the triangular system $A_1=(\begin{smallmatrix}a&c\\0&b\end{smallmatrix})$ used in Figure~\ref{fig:toy},
\begin{equation}
 A_1^k=\begin{pmatrix}
 a^k & c\sum_{j=0}^{k-1}a^{k-1-j}b^j\\
 0 & b^k
 \end{pmatrix}.
\end{equation}
The off-diagonal term is the exact lag-$k$ response from a perturbation in $y$ to a response in $x$.
For $a\neq b$ it is equivalently $c(a^k-b^k)/(a-b)$.
The full finite-horizon Green matrix is block lower triangular with blocks $A_i^{t-s}$ for $t\geq s$, so the two systems differ in source-to-destination routing at every lag even though the selected unperturbed trajectory and local eigenvalue spectrum agree.

\section{Experimental details and reduction controls}
\label{app:experimental}
\paragraph{Network and tasks.}
We use a trained 15-task leaky RNN with 256 hidden units and analyze the converged checkpoint at training step $680{,}000$ (checkpoint \texttt{snapshot\_0680000.pt}).
The task set is Pro, ReactPro, MemoryPro, Anti, ReactAnti, MemoryAnti, MemoryDecision1/2, ContextMemoryDecision1/2, MultiSensoryMemoryDecision, DMS, DNMS, DMC, and DNMC, following the flexible multitask setup of~\citet{driscoll2024flexible}.
The Green operator acts on tensors indexed by trial, time, and hidden unit and is evaluated through forward and adjoint products rather than explicit construction.

\paragraph{Task-period comparison.}
We reduce task-conditioned responses to matched task periods and compare them with the normalized Frobenius similarity in Eq.~\eqref{eq:cosine}.
On the primary network, the 16-trial exact Green comparison gives ROC-AUC $0.932$ and average precision $0.795$ over $1{,}106$ cross-task condition pairs.
In a fair eight-trial comparison, Green similarity gives AUC $0.920$ and AP $0.765$. Hidden covariance gives $0.852/0.591$, Laura-style task variance $0.941/0.812$, and a local gain-spectrum baseline $0.785/0.573$ (AUC/AP).
We therefore treat Green response as a mechanistic complement rather than a universally superior representation.
An epoch-preserving semantic-label permutation leaves a lower null AUC ($0.812\pm0.017$; observed $0.920$, $p=2\times10^{-4}$), indicating that the organization is not explained only by task-period timing.

\paragraph{Trial reliability.}
With two disjoint, coverage-matched sets of 16 trials per condition, the pairwise Green maps correlate at $0.959$, the mean matched-condition cosine is $0.953$, and the two recovery AUCs are $0.936$ and $0.920$.
Eight trials per half are noticeably less stable, so detailed pair claims use the larger stratified sample where available.

\paragraph{Temporal reductions.}
The historical temporal visualization used in Figure~\ref{fig:time} corresponds to a uniform-direction response projection, proportional to $\mathbf{1}^{\top}(\opP_{t,s}-I)\mathbf{1}/N$, and should not be interpreted as a mathematical partial trace.
We therefore check the temporal conclusion with the basis-invariant Frobenius response RMS
\begin{equation}
 E(t,s)=\sqrt{\mathbb{E}_{b}\norm{\opP^{(b)}_{t,s}-I\delta_{ts}}_F^2/N}.
\end{equation}
At step $680{,}000$, the long-range Frobenius-energy fraction is $0.017$ for ReactPro and $0.353$ for MemoryPro, with near-diagonal fractions $0.660$ and $0.167$, respectively.
At initialization, both long-range fractions are near $5\times10^{-4}$ and both near-diagonal fractions are near $0.83$.
Across all 15 tasks, persistent computations carry substantially more long-range response after training; these controls support the interpretation that training creates task-appropriate temporal routing rather than merely revealing a pathway present at initialization.

\section{Additional scope and robustness notes}
\label{app:extra}
The response representation is intentionally not presented as a universal replacement for activity or stability-based descriptions.
Cross-model comparisons show that task-level Green organization is model-dependent, while continuous-memory and stimulus-integration structure are among the most stable motifs.
A notable failure case is category memory: DMC and DNMC can have nearly identical singular-value spectra while their response directions differ substantially, illustrating why gain spectra and routing geometry answer different questions.
Likewise, task-level Green similarity correlates with gradient alignment in the primary network, but this correlation does not remain unique after controlling for task variance; we therefore do not claim that the present reduction uniquely predicts transfer or interference.
These negative results motivate the narrower descriptive claim used in the main text.

\end{document}